\def\year{2019}\relax
\documentclass[letterpaper]{article}

\usepackage{aaai19}
\usepackage{times}
\usepackage{helvet}
\usepackage{courier}
\usepackage{url}
\usepackage{amsmath}
\usepackage{amssymb}
\usepackage{graphicx}
\title{G2C: A Generator-to-Classifier Framework \\ Integrating Multi-Stained Visual Cues for Pathological Glomerulus Classification}
\author{
	Bingzhe Wu,\textsuperscript{\rm 1}\thanks{This work was done when Bingzhe Wu was a research intern at IBM Research - China.}
	Xiaolu Zhang,\textsuperscript{\rm 2}
	Shiwan Zhao,\textsuperscript{\rm 3}
	Lingxi Xie\textsuperscript{\rm 4}\\
	{\bf \Large Caihong Zeng,\textsuperscript{\rm 5}
	Zhihong Liu,\textsuperscript{\rm 5}
	Guangyu Sun\textsuperscript{\rm 1}} \\
	\textsuperscript{\rm 1}Peking University,
	\textsuperscript{\rm 2}Ant Financial Services Group,
	\textsuperscript{\rm 3}IBM Research,
	\textsuperscript{\rm 4}Johns Hopkins University\\
	\textsuperscript{\rm 5}National Clinical Research Center of Kidney Disease, Jinling Hospital\\
	wubingzhe@pku.edu.cn, yueyin.zxl@antfin.com, zhaosw@cn.ibm.com, 198808xc@gmail.com\\ zengch\_nj@hotmail.com, liuzhihong@nju.edu.cn, gsun@pku.edu.cn\\
}
 
\begin{document}
\maketitle
\begin{abstract}
Pathological glomerulus classification plays a key role in the diagnosis of nephropathy. As the difference between different subcategories is subtle, doctors often refer to slides from different staining methods to make decisions. However, creating correspondence across various stains is labor-intensive, bringing major difficulties in collecting data and training a vision-based algorithm to assist nephropathy diagnosis.

This paper provides an alternative solution for integrating multi-stained visual cues for glomerulus classification. Our approach, named {\bf generator-to-classifier} (G2C), is a two-stage framework. Given an input image from a specified stain, several {\em generators} are first applied to estimate its appearances in other staining methods, and a {\em classifier} follows to combine visual cues from different stains for prediction (whether it is pathological, or which type of pathology it has). We optimize these two stages in a joint manner. To provide a reasonable initialization, we pre-train the generators in an unlabeled reference set under an unpaired image-to-image translation task, and then fine-tune them together with the classifier.

We conduct experiments on a {\em glomerulus type classification} dataset collected by ourselves (there are no publicly available datasets for this purpose). Although joint optimization slightly harms the authenticity of the generated patches, it boosts classification performance, suggesting more effective visual cues are extracted in an automatic way. We also transfer our model to a public dataset for {\em breast cancer classification}, and outperform the state-of-the-arts significantly.
\end{abstract}
\section{Introduction}
\label{sec:introduction}

\begin{figure}
\centering
    \includegraphics[width=0.95\columnwidth]{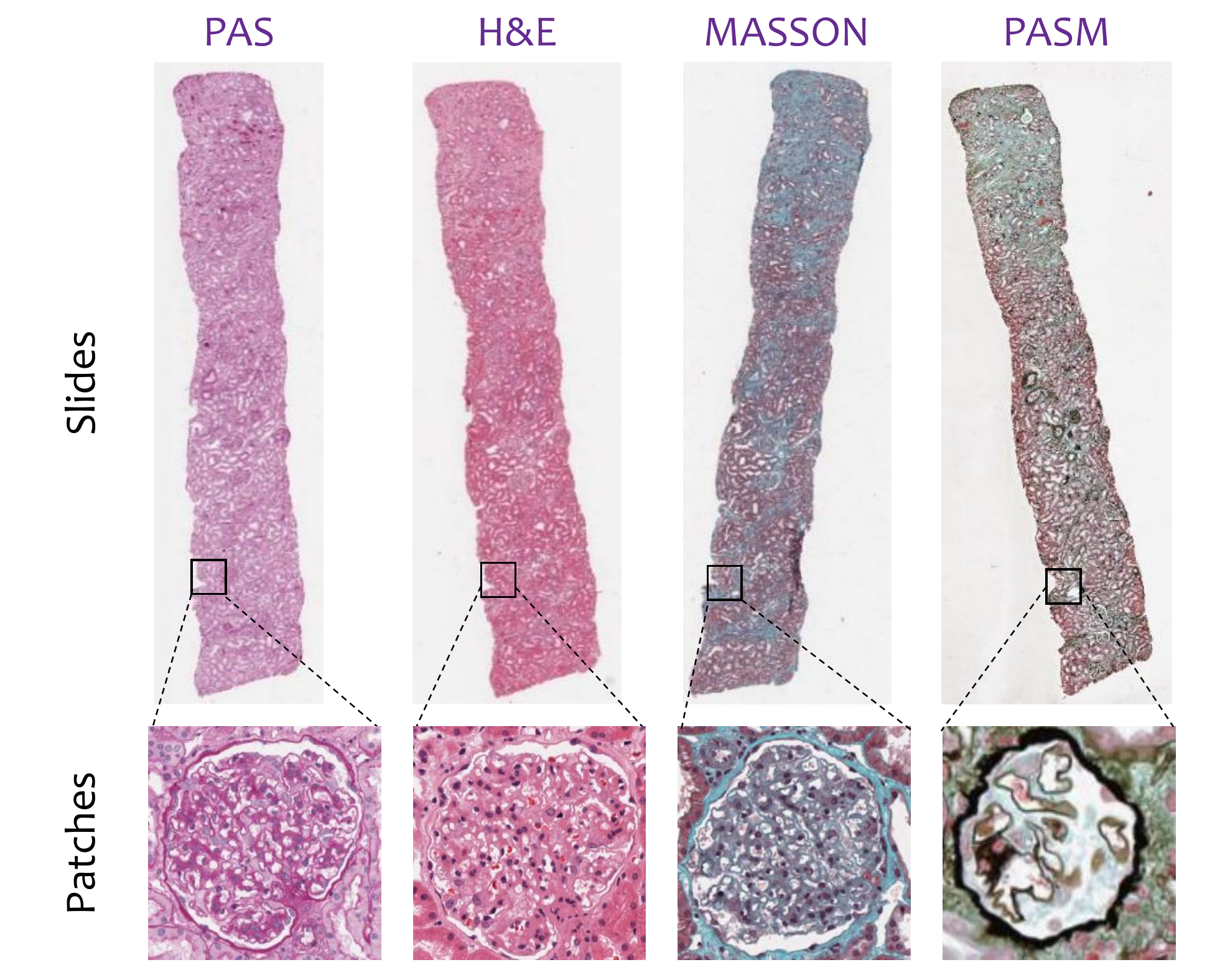}
\caption{
    Top: slides from four different staining methods ({\tt PAS}, {\tt H\&E}, {\tt MASSON} and {\tt PASM}, respectively). Bottom: four real patches, containing the {\em same} glomerulus and sampled from the {\em same} position of these slides.
}
\label{fig:example}
\end{figure}

More than $10\%$ people all over the world suffer nephropathy~\cite{Lancet2017}. An important way of diagnosis lies in a quantitative analysis of glomeruli, {\em e.g.}, discriminating between normal and abnormal samples, and further diagnosing the abnormality if necessary. In clinics, pathologists generally refer to multiple slides of the same glomerulus, generated by different staining methods, in order to collect cues from particular glomerular structures, elements, or even microorganisms to detect subtle differences among these subcategories. In this work, we consider four staining methods, namely {\tt PAS}, {\tt H\&E}, {\tt MASSON} and {\tt PASM}. As shown in Figure~\ref{fig:example}, these staining methods produce quite {\em different} appearances even for the {\em same} glomerulus.

We aim at integrating multi-stained visual cues for glomerulus classification. The main difficulty lies in the lack of annotation, {\em i.e.}, in both training and testing, labeling every glomerulus across different stains is both labor-intensive and error-prone. In our case, we are provided a partially labeled dataset on one stain ({\tt PAS}), and unlabeled data on other three stains. For most glomeruli, it is difficult to find their perfect occurrences in all four stains, thus we cannot expect a simple algorithm to learn from correspondence across different stains. This partly limits previous work~\cite{Gallego2018Glomerulus} from training classification models on multiple stains.

To this end, we propose an approach named {\bf generator-to-classifier} (G2C), with the core idea being to generate fake images in other stains to assist classification in the target stain. G2C has two stages. The first stage contains a few {\em generators}, each of which takes an input patch from one stain ({\em e.g.}, {\tt PAS}) and estimates its appearance in another stain ({\em e.g.}, {\tt H\&E}, {\tt MASSON} or {\tt PASM}). To this end, we use a popular encoder-decoder structure~\cite{zhu2017unpaired} which first down-samples the input patch to extract visual features and then up-samples to estimate its appearance in the target domain. The second stage builds a {\em classifier} upon all stains, one real and a few generated, and outputs prediction. To alleviate over-fitting, we share network weights among different branches (each branch deals with one stain), and add a cross-stain attention block after each residual block to adjust neural responses across different stains.

G2C is optimized in a joint manner so as to facilitate the collaboration between generation and classification. However, directly training everything from scratch may lead to difficulty in convergence. Therefore, we initialize each generator using CycleGAN~\cite{zhu2017unpaired}, an unpaired image-to-image translation algorithm, given weakly labeled training data, and fine-tune them with the classifier (initialized as random noise). Although this strategy may result in weaker authenticity of generated patches, it indeed enjoys higher classification accuracy, arguably because more efficient features are fed into the classifier.

We conduct experiments in two datasets for glomerulus classification and breast cancer classification, respectively. The first dataset provides a labeled image corpus in the {\tt PAS} stain, and unlabeled ones in other three stains ({\em e.g.}, {\tt H\&E}, {\tt MASSON} or {\tt PASM}). We initialize the generators using a small portion of unlabeled data (known as the reference set), and then train G2C in labeled {\tt PAS} data. G2C brings significant accuracy gain on glomerulus type classification, including distinguishing between normal and abnormal data, and discriminating two subtypes of abnormality. The second dataset only contains one ({\tt PAS}) stain, so we directly start with the fine-tuning stage, using the pre-trained generators from glomerulus data. Our approach significantly outperforms the state-of-the-arts, showing its satisfying transferability across different diseases.

We further diagnose G2C with a few comparative studies. {\bf First}, we individually analyze two stages, verifying that each generator produces high-quality patches (even professional doctors feel difficult in discriminating real and fake patches) and the classifier is an efficient solution in fusing visual information from multiple stains. {\bf Second}, joint optimization over the generators and the classifier brings a consistent accuracy gain, verifying the value of coupling information. {\bf Third}, the generation stage can be explained as an advanced way of data augmentation, which provides more constraints in other domains to alleviate over-fitting in the target domain.

In summary, the major contribution of this paper is to provide {\bf an interpretable way of adding supervision from other domains}. Compared to the recent work~\cite{shrivastava2017learning} which aimed at improving the quality of generated images, our work provides an alternative idea, {\em i.e.}, optimizing the generator with the target vision. This paper shows an example in classification, yet it has a potential of being applied to other tasks such as object detection and semantic segmentation.

The remainder of this paper is organized as follows. We first briefly review related work, and then illustrate the proposed generator-to-classifier (G2C) framework as well as the optimization method. After experimental results are shown (for glomerulus classification and breast cancer classification), we conclude this work in the final section.

\section{Related Work}
\label{sec:related_work}

Computer-aided diagnosis (CAD) plays a central role in assisting human doctors for clinical purposes. One of the most important prerequisites of CAD is medical imaging analysis, which is aimed at processing and understanding CT, MRI and ultrasound images in order to diagnose human pathology. In comparison, the digital pathology (DP) provides more accurate imaging in a small region of body tissues. Recent years have witnessed an explosion in this field, which is widely considered as one of the most promising techniques in the diagnosis, prognosis and prediction of cancer and other important diseases. This paper studies glomerulus classification from DP images. This is a key technique in diagnosing nephropathy, one of the most common types of diseases in the world.

In the conventional literatures, people made use of handcrafted features to capture discriminative patterns in digital pathology images. For example, \cite{Kakimoto2014Automated} applied an SVM on top of the Rectangular Histogram-of-Gradients (R-HOG) features for glomerulus detection, and~\cite{cruz2014automatic} designed Fuzzy Color Histogram (FCH) features~\cite{han2002fuzzy} to identify subcategories of breast cancer. Recently, the rapid development of deep learning brought more powerful and efficient solutions. Especially, as one of the most important models in deep learning, the convolutional neural networks~\cite{krizhevsky2012imagenet}\cite{simonyan2014very}\cite{szegedy2017inception} have been applied to a wide range of tasks in medical imaging analysis, including classification~\cite{Gulshan2016}~\cite{Esteva2017}~\cite{Gallego2018Glomerulus}, detection~\cite{Dou2016}, segmentation~\cite{ronneberger2015u}, {\em etc}. In the field of DP image processing, \cite{liu2017detecting} designed an automatic method to detect cancer metastases, which outperformed human doctors. \cite{chen2016mitosis} proposed a coarse-to-fine approach for mitosis detection. In~\cite{Janowczyk2016Deep}, a unified framework was presented for a series of tasks, including nuclei segmentation and mitosis detection. Our work is closely related to~\cite{Pedraza2017Glomerulus}, which trained deep networks to outperform handcrafted features in glomerulus classification.

The importance of using multiple stains for digital pathology diagnosis is emphasized by the doctors in our team. However, annotating data correspondence is difficult and time-consuming, so we turn to the family of Generative Adversarial Networks (GANs)~\cite{goodfellow2014generative} to perform image-to-image translation. There are generally two types of translation algorithms, paired~\cite{isola2016image} and unpaired~\cite{kim2017learning}\cite{liu2017unsupervised}\cite{yi2017dualgan}\cite{zhu2017unpaired}, which differ from each other in the organization of input data. The former type is often more accurate, while the latter type can be used in the scenario of missing data correspondence, which fits the requirement of this work.

\begin{figure*}[!t]
\centering
    \includegraphics[width=1.95\columnwidth]{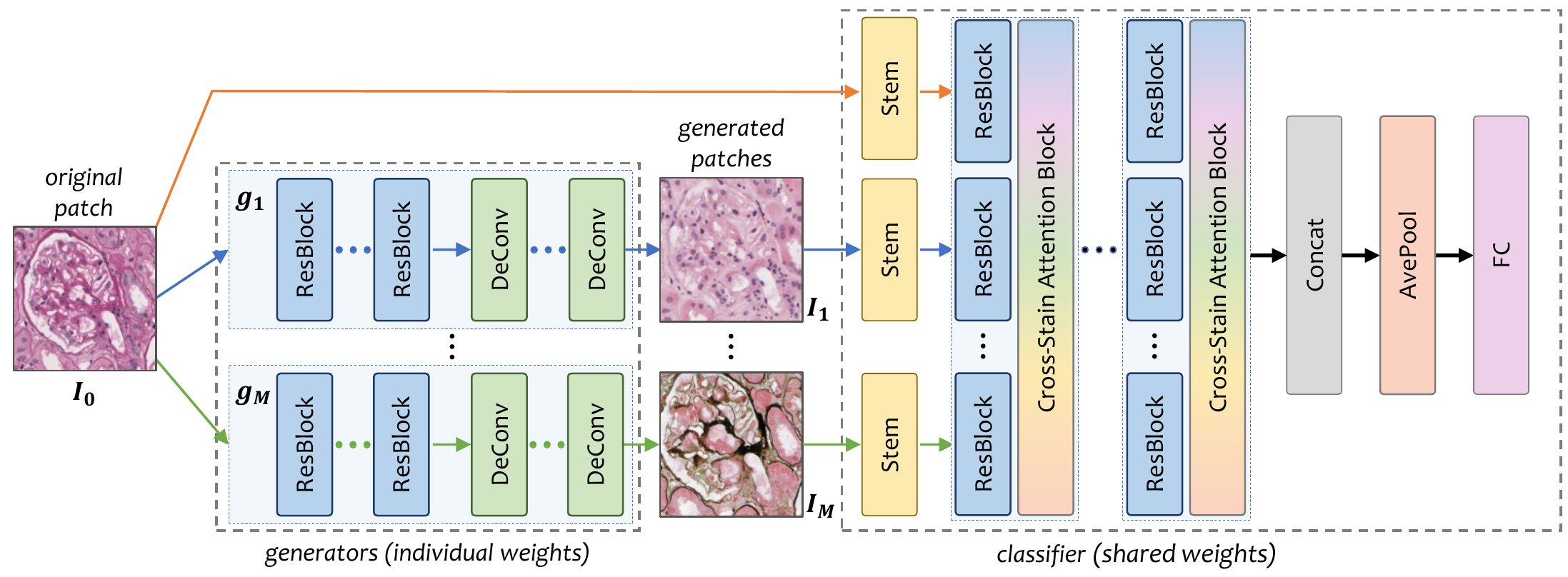}
\caption{
    The overall {\bf generator-to-classifier} (G2C) framework. The left part illustrates the $M$ {\em generators}, and the right part the {\em classifier}, in which all $M+1$ branches share the same weights. When an input patch comes, $m$ other stains are generated, and then combined with the original one for classification. The entire framework is end-to-end, and all the modules can be optimized in a joint manner.
}
\label{fig:framework}
\end{figure*}

\section{Our Approach}
\label{sec:approach}

\subsection{Backgrounds}
\label{sec:approach:backgrounds}

Staining is a popular way to highlight the important features of a soft tissue. Each staining method has both advantages and disadvantages~\cite{fogo2014fundamentals}. For example, the {\tt PAS} stains glomerular basement membranes, mesangial matrix and tubular basement membranes red (positive), while the {\tt PASM} colors the same component black, providing a clear contrast between positively and negatively staining structures. Integrating multi-stained information is very important for pathology image analysis, {\em e.g.}, for clinical purposes.

However, in collecting a large dataset for glomerulus classification, it is difficult to label each glomerulus under all staining methods, because (i) finding correspondence between stains is labor-intensive, and (ii) only a small portion of glomeruli can be clearly seen in multiple stains\footnote{Each slide in digital pathology can be stained only once. Even if a set of neighboring slides containing the same glomerulus are used in various stains, its appearance may not be identical due to the difference in slicing positions.}. Therefore, we set our goal to be {\em glomerulus classification from single-stained inputs}. To be specific, each input patch contains a glomerulus from the {\tt PAS} stain. Meanwhile, a small corpus of $100$ unlabeled patches is also provided for each stain (including the {\tt PAS} stain and other three stains). These four corpora form the {\em reference set} used for initializing cross-stain generators.

\subsection{Formulation}
\label{sec:approach:formulation}

Let the input be a patch $\mathbf{I}$ sampled from a slide with {\tt PAS} staining. The goal is to design a model $\mathbb{M}: {t}=\mathbf{f}\!\left(\mathbf{I};\boldsymbol{\theta}\right)$, where $t$ is the class label, and $\boldsymbol{\theta}$ are the model parameters, {\em e.g.}, the learnable weights in a convolutional neural network.

Recall that our goal is to start with one stain, generate fake images for other stains, and finally make prediction based on all stains. We formulate the above flowchart into a joint optimization task, in which a few {\em generators} are first used to generate other stains ({\tt H\&E}, {\tt MASSON} and {\tt PASM}) from the input {\tt PAS} stain, and a {\em classifier} follows to extract features from all these images and outputs the final prediction. Following this, we decompose the function $\mathbf{f}\!\left(\cdot\right)$ into two stages taking charge of generation and classification, respectively. The overall flowchart is illustrated in Figure~\ref{fig:framework}.

\noindent
$\bullet$\quad{\bf The Generation Network}

The first set of modules, named {\em generators}, play the role of generating patches of different staining methods from the input patch $\mathbf{I}$. We denote the generated patch set by ${\mathcal{I}}={\left\{\mathbf{I}_0,\mathbf{I}_1,\ldots,\mathbf{I}_M\right\}}$, in which ${\mathbf{I}_0}\doteq{\mathbf{I}}$ is the source patch, and all other $M$ ones are generated using a parameterized model ${\mathbf{I}_m}={\mathbf{g}_m\!\left(\mathbf{I}_0;\boldsymbol{\theta}_m^\mathrm{G}\right)}$, ${m}={1,2,\ldots,M}$.

Each generator consists of several down-sampling units and the same number of up-sampling units. As the number of residual blocks increases, the classification accuracy goes up and gradually saturates. In practice, As a tradeoff between accuracy and efficiency, we use $6$ resolution-preserved residual blocks, $2$ convolutional layers (kernel size is $3$, stride is $2$) for down-sampling, and $2$ deconvolution layers (kernel size is $4$, stride is $2$) for up-sampling. Following~\cite{zhu2017unpaired}, each convolutional layer is followed by an instance normalization layer.

\noindent
$\bullet$\quad{\bf The Classification Network}

The second module is named a {\em classifier}, which integrates information from all patches (one real and $M$ generated) for classification. We denote this stage as ${t}={\mathbf{c}\!\left(\mathbf{I}_0,\mathbf{I}_1,\ldots,\mathbf{I}_M;\boldsymbol{\theta}^\mathrm{C}\right)}$.

Conceptually, the parameters $\boldsymbol{\theta}^\mathrm{C}$ need to capture visual properties from all stains. One choice is to train $M+1$ sub-networks with parameters $\boldsymbol{\theta}_0^\mathrm{C},\boldsymbol{\theta}_1^\mathrm{C},\ldots,\boldsymbol{\theta}_0^\mathrm{C}$, respectively. Suppose the number of parameters for each sub-network is $O\!\left(L\right)$, this strategy would contribute $O\!\left(ML\right)$ parameters to the entire model. Another choice would be using the same set of parameters $\boldsymbol{\theta}^\mathrm{C}$ in each branch (this is reasonable as different stains share similar visual features), but allowing several {\em cross-stain attention blocks} to swap information among different branches. As we shall detail below, these blocks are often equipped with a small amount of parameters, and, consequently, the number of parameters is reduced to $O\!\left(ML'+L\right)$, in which $O\!\left(L'\right)$ is the number of parameters of a cross-stain attention block and ${L'}\ll{L}$. This reduces the risk of over-fitting especially for small datasets.

Following this idea, the designed classifier is a multi-path model consisting of $M+1$ branches, each of which is a variant of the deep residual network~\cite{he2016deep}. A stem block~\cite{szegedy2017inception} is used to replace the original $7\times7$ convolutional layer\footnote{Experiments show that using the stem block consistently improves classification accuracy by more than $1\%$.}, followed by a few down-sampling units ($3$ residual blocks and a stride-$2$ pooling layer). A cross-stain attention block follows each residual block, in which we follow~\cite{hu2018senet} to first down-sample neural responses from all stains (squeeze), then pass it throught two fully-connected layers, and multiply it to each channel of the original responses (excitation). Lastly, $M+1$ feature vectors are concatenated, average-pooled and fully-connected to the final prediction layer.

In summary, the overall framework is a composed function of the generator and the classifier, {\em i.e.},
\begin{equation}
\label{eqn:overall}
{t}={\mathbf{f}\!\left(\mathbf{I};\boldsymbol{\theta}\right)}\doteq{\mathbf{c}\circ\mathbf{g}\!\left(\mathbf{I};\left\{\boldsymbol{\theta}_m^\mathrm{G}\right\}_{m=1}^M,\boldsymbol{\theta}^\mathrm{C}\right)}.
\end{equation}
Note that when ${M}={0}$, our model degenerates to that using one single stain for classification. Sharing parameters over $M+1$ branches enables us to fairly compare our model and the baseline {\em at the classification stage}. This idea also originates from the doctors in our team, who suggests that different staining images provide complementary information in diagnosis, but the basic principles to recognize them should remain unchanged.

\subsection{Optimization}
\label{sec:approach:optimization}

We hope to jointly optimize Eqn~\eqref{eqn:overall} so as to enable the parameters of the generators and the classifier to collaborate with each other. But, according to our motivation, the generator should be able to produce some reasonable images corresponding to other staining methods. Therefore, we suggest a two-stage training process, in which we first train the generative networks using some unlabeled data covering different stains, and then fine-tune the generator together with the classifier towards higher recognition accuracy.

\noindent
$\bullet$\quad{\bf Initializing the Generators}

\noindent
Due to the lack of data correspondence, the generators are initialized by a task known as unpaired image-to-image translation. $100$ patches from the source stain and another $100$ from the target stain are provided. Note that all these patches are unlabeled, and may even not contain glomeruli. We use a recent approach named CycleGAN~\cite{zhu2017unpaired}, which trains a reverse generator, denoted by $\widehat{\mathbf{g}}_m$, to translate the patches generated by $\mathbf{g}_m$ back to the source stain. $\widehat{\mathbf{g}}_m$ shares the same structure with $\mathbf{g}_m$. We follow the original implementation in setting hyper-parameters.

Note that, if additional annotations on the target domains are available, we can use more accurate image-to-image translation algorithms such as~\cite{isola2016image} to initialize our model. In a high-level perspective, this initialization process also eases the training stage by providing mid-level supervision, forcing the generator to produce reasonable patches, and reducing the risk of over-fitting a very deep network to a limited amount of training data.

\noindent
$\bullet$\quad{\bf Fine-tuning Generators with the Classifier}

\noindent
In this stage, we train the classifier together with the generators in a fully-supervised manner (each glomerulus is assigned a class label). Our goal is no longer high-quality generation, but accurate classification. Therefore, the reference set containing multi-stained, unlabeled data is not used, and all the generators $\widehat{\mathbf{g}}_m$, ${m}={1,\ldots,M}$, are simply discarded.

In network training, we use Stochastic Gradient Descent with a momentum of $0.5$. We perform a total of $30$ epochs. The learning rate is set to be $0.01$ at the beginning, and divided by $10$ after every $5$ epochs. In the first $5$ epochs, we freeze all parameters in the generators in order to initialize the classifier with fixed generated samples, so as to improve the stability in training. Note that freezing the generators in all $30$ epochs leads to training the generators and the classifier individually. We shall show in experiments that joint optimization leads to significant accuracy gain over all our generator-to-classifier models.

As we shall see in Table~\ref{tab:dataset}, the numbers of training data in different classes may be imbalanced. To prevent a category with fewer training samples from being swallowed by another category, we introduce the focal loss function~\cite{lin2017focal} which brings slight but consistent accuracy gain in each individual classification task.

\subsection{Why Our Approach Works?}
\label{sec:approach:analysis}

It remains to discuss why our algorithm works well (see the next section for quantitative results). The key contribution naturally comes from the ability of simulating different staining methods, and enabling jointly optimization so that the classifier takes advantage of complementary information. To confirm that these information comes from the {\em authenticity} of the generators, we perform a user study on the professional doctors in our team, and verify that it is even difficult for them to distinguish the generated patches from real ones.

Moreover, we make some comments on another question: as all the information comes from the input image ({\em i.e.}, either in the baseline or our algorithm, the classifier sees the same amount of information), what is brought into the system that leads to the accuracy gain? We explain this to be a guided way of extracting high-quality features. Note that in training a glomerulus classifier, the amount of data is most often very limited. Our dataset merely contains $2\rm{,}650$ cases for {\em ss}-vs-{\em gs} classification, with less than $1\rm{,}000$ training images in the {\em ss} subtype. When a powerful classifier, say a very deep neural network, is used, the training data can be explained in a lot of different ways, but most of them do not learn from human knowledge, and thus do not fit the testing data very well. Our algorithm, by introducing the knowledge from human doctors that other staining methods are helpful to classification, forces the model to rely a great deal on multi-stained data. We believe this algorithm to endure fewer risks especially in the scenario of limited data. This is verified by investigating the over-fitting issue, shown in the diagnostic part, and transferring our models to another dataset for breast cancer classification, shown in the last part of our experiments.

Last but not least, although our approach can be explained as an advanced way of data augmentation, it introduces a complementary prior (with the help of an unlabeled reference set) to conventional data augmentation (assuming that semantics of a patch remains unchanged when it is flipped, cropped, {\em etc.}). In experiments, we find that (1) our approach achieves much more accuracy gain than data augmentation, and (2) these two methods can be used together towards the best performance.

\section{Experimental Results}
\label{sec:experiments}

\subsection{Dataset and Settings}
\label{sec:experiments:settings}

\begin{table}
\centering
\begin{tabular}{|l||r|r|r||r|}
\hline
{}       & {\em ss} & {\em gs}     & {\em noa}    & Total         \\
\hline\hline
Training &    $648$ & $2\rm{,}002$ & $7\rm{,}000$ & $ 9\rm{,}650$ \\
\hline
Testing  &    $237$ &        $618$ & $2\rm{,}828$ & $ 3\rm{,}683$ \\
\hline\hline
Total    &    $885$ & $2\rm{,}620$ & $9\rm{,}828$ & $13\rm{,}333$ \\
\hline
\end{tabular}
\caption{
    Number of annotated glomeruli of each subcategory in the {\tt PAS} stain. The {\em ss} and {\em gs} categories compose the high-level abnormal category, denoted by {\em s}.
}
\label{tab:dataset}
\end{table}

We collect a dataset for glomerulus classification. As far as we know, there is no public dataset for this purpose (existing ones~\cite{Pedraza2017Glomerulus} worked on glomerulus-vs-non-glomerulus classification). Our dataset is collected from $209$ patients, each of which has several slides from four staining methods, namely {\tt PAS}, {\tt H\&E}, {\tt MASSON} and {\tt PASM}. In all {\tt PAS} slides, we ask the doctors to manually label a bounding box for each glomerulus, and annotate its subcategory. The doctors annotate with confidence, {\em i.e.}, only those {\tt PAS} patches containing enough information to make decisions are preserved. The subcategories include {\em global sclerosis} ({\em gs}), {\em segmental sclerosis} ({\em ss}), and being normal ({\em none of the above} or {\em noa}). Global sclerosis and segmental sclerosis are two levels of {\em glomerulosclerosis} (denoted by {\em s}). Advised by the doctors, we consider two classification tasks, dealing with {\em s}-vs-{\em noa} and {\em gs}-vs-{\em ss}, respectively. The first task is aimed at discriminating abnormal glomeruli from normal ones, and the second task goes one step further to determine abnormality of the abnormal glomeruli. To deal with imbalanced label distribution (see Table~\ref{tab:dataset}), we report category-averaged accuracies~\cite{brodersen2010balanced} in the following experiments.

All $209$ patients in the dataset are split into a training set ($149$ patients) and a testing set ($60$ patients). There are in total $9\rm{,}650$ annotated patches (each contains one glomerulus) in the training set and $3\rm{,}683$ in the testing set. The statistics of all subcategories are provided in Table~\ref{tab:dataset}. To initialize the generators, we construct a reference set for each of the other stains by randomly cropping $100$ patches from the unlabeled data. These patches may not contain a glomerulus, or contain a part of it, but as we shall see later, such weakly-labeled data are enough to train the generative networks.

Setting ${M}={0}$ leads to our baseline model in which only the {\tt PAS} stain is used for classification. We denote it by {\tt PAS\_ONLY}. We also provide several competitors, which differ from each other in the type(s) of stains generated to assist classification. These variants are denoted by {\tt PAS\_H\&E}, {\tt PAS\_MASSON}, {\tt PAS\_PASM} and {\tt PAS\_ALL}, respectively. Among which, {\tt PAS\_ALL} integrates information from all the other three stains, {\em i.e.}, ${M}={3}$.

In our dataset, there are much fewer abnormal glomerulus patches than normal ones. To prevent over-fitting, we perform data augmentation by (i) randomly flipping input patches vertically and/or horizontally, and (ii) performing random color jittering, including changing the brightness and saturation of input patches. All input patches are rescaled into $224\times224$, and pixel intensity values are normalized into $\left[0,1\right]$.

\subsection{Quantitative Results}
\label{sec:experiments:results}

\begin{table}
\centering
\begin{tabular}{|l||r|r||r|}
\hline
{}                & {\em s}          & {\em noa}        & Average          \\
\hline\hline
{\tt PAS\_ONLY}   & $90.76$          & $90.73$          & $90.74$          \\
\hline
{\tt PAS\_H\&E}   & $92.74$          & $92.36$          & $92.55$          \\
\hline
{\tt PAS\_MASSON} & $92.39$          & $91.72$          & $92.06$          \\
\hline 
{\tt PAS\_PASM}   & $93.09$          & $91.83$          & $92.46$          \\
\hline
{\tt PAS\_ALL}    & $\mathbf{93.68}$ & $\mathbf{92.99}$ & $\mathbf{93.34}$ \\
\hline
{\tt PAS\_ALL+}   & $\mathbf{94.15}$ & $\mathbf{93.02}$ & $\mathbf{93.68}$ \\
\hline
\end{tabular}
\caption{
    Category-wise and averaged classification accuracies ($\%$) in the {\em s}-vs-{\em noa} task. {\tt PAS\_ALL+} indicates that cross-stain attention is added for feature re-weighting.
}
\label{tab:s_noa}
\end{table}

\subsubsection{Level-1 Classification: {\em s}-vs-{\em noa}}

We first evaluate classification accuracy in discriminating abnormal glomeruli (denoted by {\em s}) from normal ones (denoted by {\em noa}). Results are summarized in Table~\ref{tab:s_noa}. One can observe that introducing additional stain(s) consistently improves classification accuracy. An interesting but expected phenomenon emerges by looking into category-wise accuracies. For example, based on the {\tt PAS} stain, adding {\tt H\&E} produces a higher classification rate in the normal ({\em noa}) category, while {\tt MASSON} works better in finding abnormal ({\em s}) glomeruli. This suggests that different stains provide complementary information to assist diagnosis, and verifies the motivation of this work. Therefore, it is not surprising that combining all other stains obtains consistent accuracy gain over other competitors. In particular, the {\tt PAS\_ALL} model outperforms the {\tt PAS\_ONLY} model by $2.60\%$ in the averaged accuracy, or a $28.08\%$ relative drop in classification error. Our best model is {\tt PAS\_ALL+}, which adds cross-stain attention and further improves classification accuracy. We will analyze the benefit of this module in the diagnostic part.

\begin{table}
\centering
\begin{tabular}{|l||r|r||r|}
\hline
{}                & {\em ss}         & {\em gs}         & Average          \\
\hline\hline
{\tt PAS\_ONLY}   & $78.05$          & $96.76$          & $87.41$          \\
\hline
{\tt PAS\_H\&E}   & $79.23$          & $96.87$          & $88.05$          \\
\hline
{\tt PAS\_MASSON} & $78.31$          & $96.79$          & $87.55$          \\
\hline
{\tt PAS\_PASM}   & $81.43$          & $97.57$          & $89.50$          \\
\hline
{\tt PAS\_ALL}    & $\mathbf{81.59}$ & $\mathbf{98.20}$ & $\mathbf{89.90}$ \\
\hline
{\tt PAS\_ALL+}   & $\mathbf{82.23}$ & $\mathbf{98.67}$ & $\mathbf{90.45}$ \\
\hline
\end{tabular}
\caption{
    Category-wise and averaged classification accuracies ($\%$) in the {\em ss}-vs-{\em gs} task. {\tt PAS\_ALL+} indicates that cross-stain attention is added for feature re-weighting.
}
\label{table:ss_gs}
\end{table}

\subsubsection{Level-2 Classification: {\em ss}-vs-{\em gs}}

Next, we further categorize the abnormal ({\em s}) glomeruli into two subtypes, namely, investigating the {\em ss}-vs-{\em gs} classification task. Advised by the doctors in our team, we only consider those correctly categorized abnormal patches in Level-1. Results are summarized in Table~\ref{table:ss_gs}. Qualitative analysis gives similar conclusions, {\em i.e.}, different stains provide complementary information, therefore it is instructive to combine all stains for accurate classification. It is worth noting that in these two abnormal subcategories, segmental sclerosis ({\em ss}) suffers lower accuracy compared to global sclerosis ({\em gs}), which is partly caused by the limited amount and imbalance of training data. This is alleviated by incorporating generated patches from other stains as augmented data. Compared to {\tt PAS\_ONLY}, the {\tt PAS\_ALL} model significantly improves the {\em ss} classification accuracy by $3.54\%$, and the overall accuracy by $2.49\%$ (a $19.78\%$ relative drop in classification error). Similarly, {\tt PAS\_ALL+} benefits from cross-stain attention and goes one step beyond equally weighting all stains ({\em i.e.}, {\tt PAS\_ALL}).

%We have performed a statistical test on the two tasks, which reports the p-value of $0.011$ and $0.0014$ in the {\em ss}-vs-{\em gs} task and  {\em s}-vs-{\em noa} task, respectively. Both cases show statistical significance.
We perform statistical tests on the two tasks, reporting that {\tt PAS\_ALL} outperforms {\tt PAS\_ONLY} significantly with p-value of $0.011$ and $0.0014$ in the {\em s}-vs-{\em noa} task and {\em ss}-vs-{\em gs} task, respectively. 
\subsection{Discussions}
\label{sec:experiments:discussions}

This part provides several discussions on our approach. First we observe the performance of two stages (generation and classification) individually, and then we discuss the benefit of joint optimization and how our approach helps to alleviate over-fitting especially in small datasets. We also show a comparison with conventional data augmentation strategies. 

\noindent
$\bullet$\quad{\bf Qualitative Studies on the Generators}

\begin{figure}[t]
\centering
    \includegraphics[width=0.95\columnwidth]{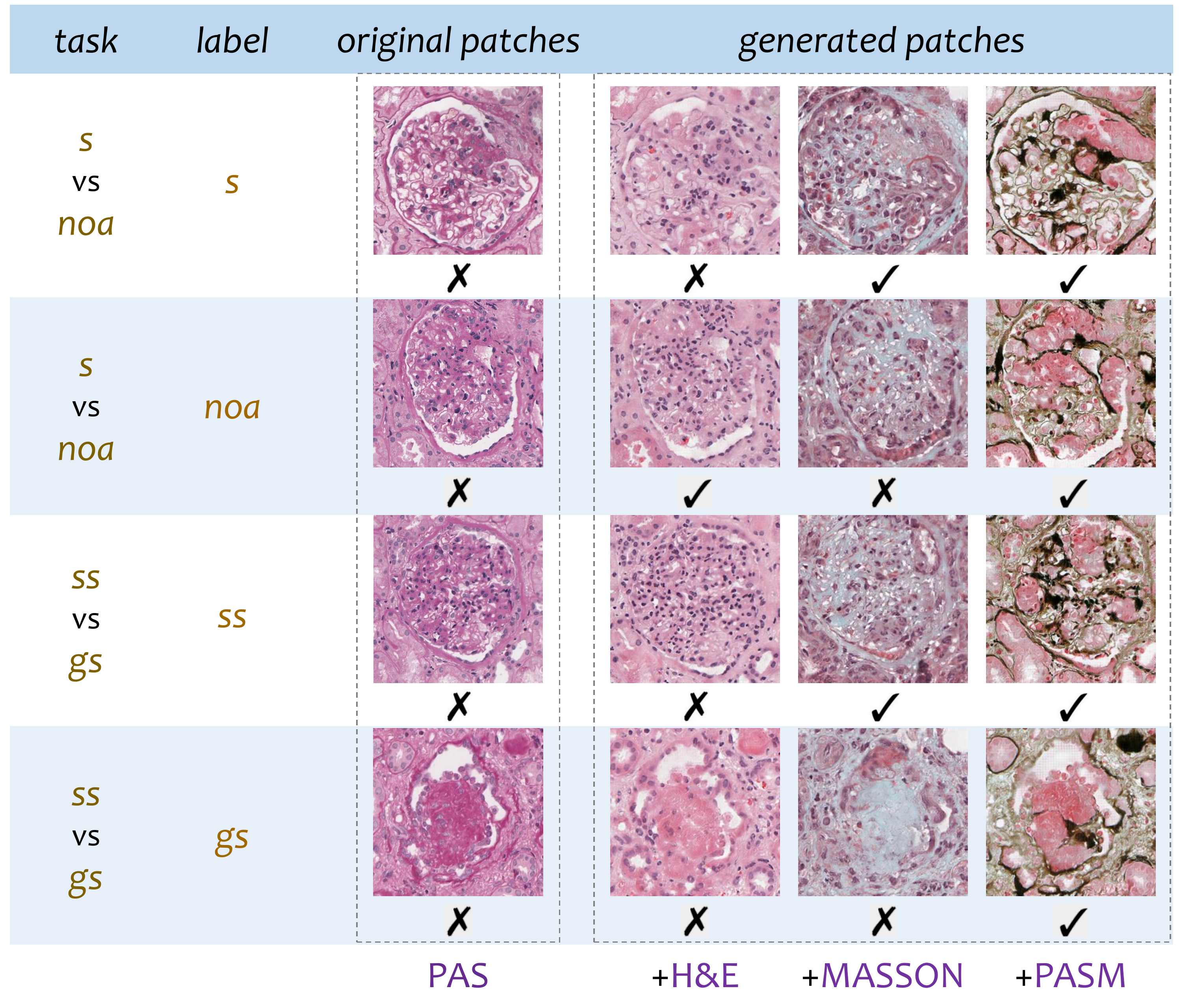}
\caption{
    Different stains provide complementary information to assist glomerulus classification. In each row, the original patch is mis-classified using the {\tt PAS} stain alone (marked by a cross), but turned into correctness after integrating some of other generated stains (marked by a tick). {\bf All these glomeruli are correctly classified using all four stains ({\em i.e.}, {\tt PAS\_ALL} and {\tt PAS\_ALL+}).}
}
\label{fig:visualization}
\end{figure}

\noindent
To confirm the authenticity of the generated patches, we perform a study by asking the doctors in our team to discriminate the generated patches from the real ones. We sample $50$ patches from all the generated ones, combine them with $50$ real patches, show them one-by-one to the pathologists and record their judgments. The average accuracy over three pathologists is $70.0\%$ (random guess reports $50\%$), suggesting an acceptable quality to professional doctors.

Figure~\ref{fig:visualization} shows several examples in which glomeruli are misclassified using the {\tt PAS} stain alone, and rescued by the generated stains. We note that each failure case in {\tt PAS} can be helped by one or a few other stains. In clinics, these generated patches may also assist doctors in case that a {\tt PAS} patch does not contain sufficient information.

\noindent
$\bullet$\quad{\bf The Design of Classifier}

\noindent

To reduce the number of parameters, we share parameters among different branches of the classifier, which is based on the assumption that visual features extracted from different stains are mostly similar. The rationality of this assumption is verified by the results in Tables~\ref{tab:s_noa} and~\ref{table:ss_gs}. Moreover, adding cross-stain attention blocks consistently boosts performance of a multi-branch classifier, {\em e.g.}, the overall error drop is $0.34\%$ ($5.10\%$ relatively) and $0.55\%$ ($5.44\%$ relatively) for two tasks, respectively. Note that this is achieved by adding merely $5.27\%$ more parameters to the classifier.

We also evaluate the use of cross-stain attention blocks in the scenario of fewer branches, and observe smaller improvement, {\em e.g.}, on top of {\tt PAS\_ONLY}, the overall accuracy gain is $0.30\%$ and $0.09\%$, respectively. This suggests the existence of cross-domain feature difference, yet a light-weighted module is sufficient in dealing with it.

\noindent
$\bullet$\quad{\bf The Benefit of Joint Optimization}

\noindent
In addition, joint optimization brings significant gain in classification accuracy. In comparison to the model in which the generators and the classifier are optimized individually ({\em i.e.}, the weights of the generators are frozen throughout the fine-tuning stage), the jointly optimized models ({\tt PAS\_ALL}) achieve $1.10\%$ and $1.54\%$ boosts on {\em s}-vs-{\em noa} and {\em ss}-vs-{\em gs} classification, respectively. In particular, the error of the most challenging {\em ss} class is reduced from $20.68\%$ to $18.41\%$ ($2.27\%$ absolute or $10.98\%$ relative drop).

\noindent
$\bullet$\quad{\bf The Over-fitting Issue}

\begin{table}
\centering
\begin{tabular}{|l||r|r||r|}
\hline
{}                & Training & Testing & Gap             \\
\hline\hline
{\tt PAS\_ONLY}   & $96.48$  & $87.41$ & $9.07$          \\
\hline
{\tt PAS\_H\&E}   & $94.90$  & $88.05$ & $6.85$          \\
\hline
{\tt PAS\_MASSON} & $93.23$  & $87.55$ & $5.68$          \\
\hline
{\tt PAS\_PASM}   & $95.03$  & $89.50$ & $5.53$          \\
\hline
{\tt PAS\_ALL}    & $94.87$  & $89.90$ & $4.97$          \\
\hline
{\tt PAS\_ALL+}   & $95.07$  & $90.45$ & $\mathbf{4.62}$ \\
\hline
\end{tabular}
\caption{
    For different models on {\em ss}-vs-{\em gs} classification, we report training and testing accuracy as well as the gap between them. {\tt PAS\_ALL+} indicates that cross-stain attention is added for feature re-weighting.
}
\label{tab:overfitting}
\end{table}

\noindent
In the area of medical imaging analysis, recognition accuracy is often limited by the insufficiency of training data. Although being significantly larger than any publicly available glomerulus datasets, there are less than $1\rm{,}000$ training samples for the {\em ss} subcategory. Considerable over-fitting may arise because the testing set contains some cases which are not covered by the training set.

Generating patches in other stains alleviates over-fitting to some extents. It provides complementary information to geometry-based data augmentation such as flip and rotation, as the generators bring in some {\em priors} learned from the reference sets ($100\times3$ unlabeled patches from other stains), forcing the training data to be explained in a more reasonable manner. To verify this, we record both training and testing accuracies for each of the five models for {\em ss}-vs-{\em gs} classification in Table~\ref{tab:overfitting}. Using multiple stains leads to a higher testing accuracy but a lower training accuracy, which is the consequence of a stronger constraint (multiple stains need to be explained collaboratively) in training deep neural networks.

\noindent
$\bullet$\quad{\bf Comparison with data augmentation}

To compare with conventional data augmentation methods, we conduct a comparative experiment on the {\em ss}-vs-{\em gs} task. In Table~\ref{table:ss_gs}, we apply data augmentation to {\tt PAS\_ONLY}, while all the other models are trained without data augmentation. We also train the {\tt PAS\_ONLY} without data augmentation and get an accuracy at $86.21\%$. 
As shown in Table~\ref{table:ss_gs}, the accuracies of {\tt PAS\_ONLY} with data augmentation and {\tt PAS\_ALL+} are $87.41\%$ and $90.45\%$, respectively. Our method {\tt PAS\_ALL+} improves the accuracy by $4.24\%$ while the data augmentation improves the accuracy by $1.20\%$, which indicates that our method outperforms conventional data augmentation. 

\subsection{Transferring to Breast Cancer Classification}
\label{sec:experiments:transferring}

%\defcitealias{Janowczyk2016Deep}{Janowczyk {\em et al.}, 2016}
\begin{table}
\centering
\begin{tabular}{|l||r|r|}
\hline
{}                        & F1-score         & Accuracy ($\%$)  \\
\hline\hline
\cite{han2002fuzzy}      & $0.675$          & $78.7 $          \\
\hline
\cite{cruz2014automatic} & $0.718$          & $84.23$          \\
\hline
(Janowczyk~{et~al.} 2016) & $0.765$          & $84.68$          \\
\hline
Ours ({\tt PAS\_ALL+})    & $\mathbf{0.841}$ & $\mathbf{88.28}$ \\
\hline
\end{tabular}
\caption{
    Comparison of F1-scores and balanced accuracies on the breast cancer classification task. [Janowczyk~{\em et~al.}, 2016] is the baseline ({\em i.e.}, {\tt PAS\_ONLY}).}
\label{table:transferring_results}
\end{table}
To further demonstrate the effectiveness of our approach,
we apply it to a publicly available dataset for invasive ductal carcinoma (IDC) classification\footnote{{\tt http://www.andrewjanowczyk.com/}}, which contains $277\rm{,}524$ patches of $50\times50$ pixels ($198\rm{,}738$ IDC-negative and $78\rm{,}786$ IDC-positive). 
To make a fair comparison, we reproduce~\cite{Janowczyk2016Deep} with the same network architecture (AlexNet) on the {\tt PAS} stain alone~(baseline model). As all patches in this dataset are {\tt PAS}-stained, we do not train new generators from scratch, but simply transfer the pre-trained ones from our dataset, and fine-tune them with the new classifier. We apply our best configuration learned from the previous task, namely, using all four stains and adding cross-stain attentions. This model is denoted by {\tt PAS\_ALL+}.

\begin{figure}[t]
\centering
    \includegraphics[width=0.85\columnwidth]{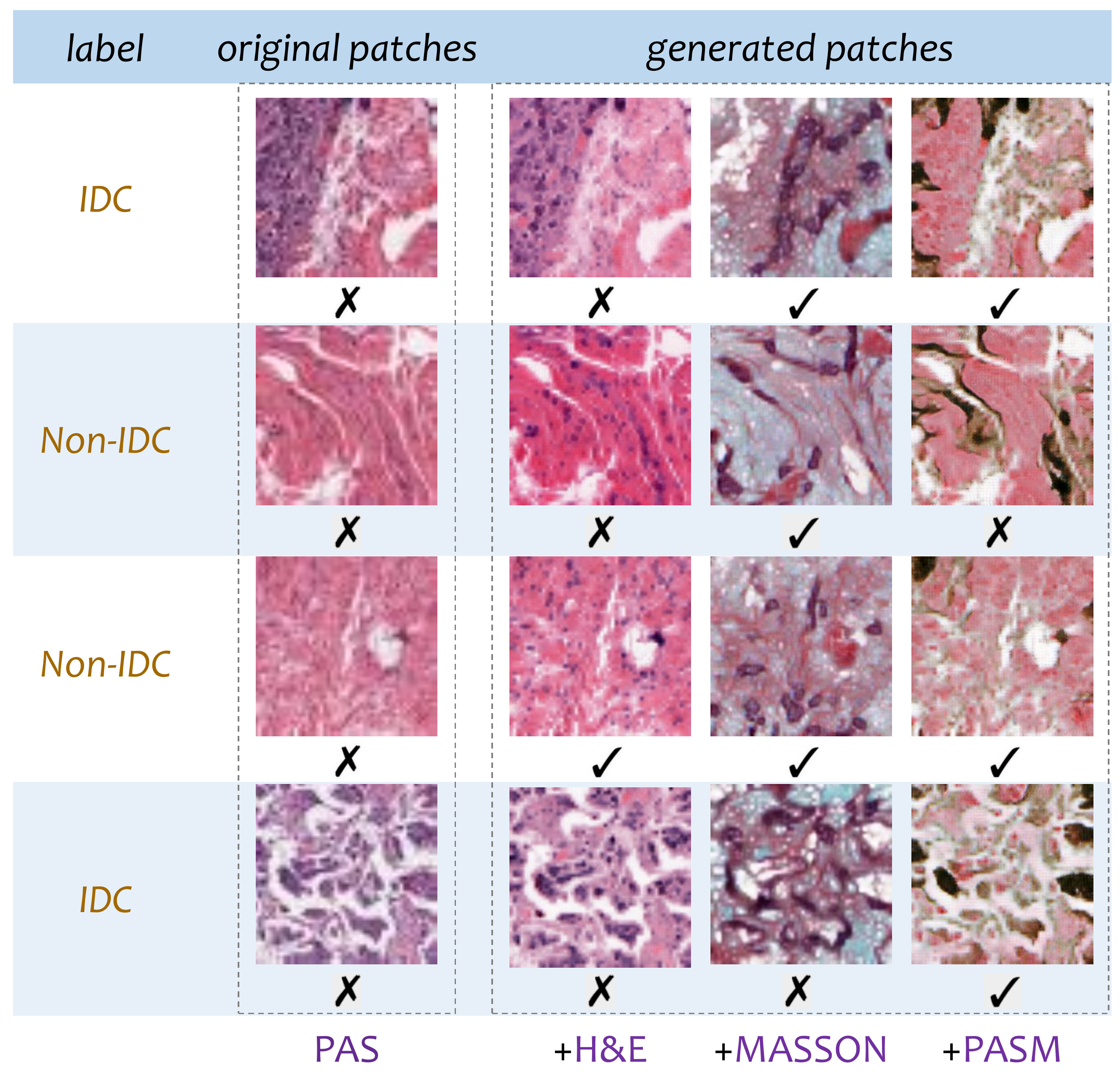}
\caption{
    Different stains provide complementary information to assist breast cancer classification. In each row, the original patch is mis-classified using the {\tt PAS} stain alone (marked by a cross), but turned into correctness after integrating some of other generated stains (marked by a tick). {\bf All these patches are correctly classified using all four stains ({\em i.e.}, {\tt PAS\_ALL} and {\tt PAS\_ALL+}).}
}
\label{fig:visualization2}
\end{figure}

Results are shown in Table~\ref{table:transferring_results}. In terms of both F1-score and classification accuracy, our approach significantly outperforms~\cite{Janowczyk2016Deep}, as well as two previous methods with handcrafted features~\cite{han2002fuzzy} and relatively shallow CNNs~\cite{cruz2014automatic}. Similarly, we visualize some examples in Figure~\ref{fig:visualization2}.

Hence, we conclude on the effectiveness of our training strategy. The first stage, {\em i.e.}, initializing the generators, can be performed in a fixed reference set ({\em e.g.}, containing glomeruli); when another dataset is available, we can directly move on to the second stage, {\em i.e.}, fine-tuning a new classifier with these generators.

\section{Conclusions}
\label{sec:conclusions}

In this paper, we present a novel approach for glomerulus classification in digital pathology images. Motivated by the need of generating multiple stains for accurate diagnosis, we design a {\bf generator-to-classifier} (G2C) network, and perform an effective two-stage training strategy. {\bf The key innovation lies in the mechanism which enables several generators and a classifier to collaborate in both training and testing.} A large dataset is collected by the doctors in our team, which is much larger than any publicly available ones. Our approach achieves considerably higher accuracies over the baseline, and transfers reasonably well to another digital pathology dataset for breast cancer classification.

This research paves a new way of data enhancement in medical imaging analysis, which is more advanced complementary to conventional data augmentation. Transferring this idea to other types of data generation, {\em e.g.}, integrating CT scans from the {\em arterial phase} and the {\em venous phase} for organ segmentation, is promising and implies a wide range of clinical applications.
\section{Acknowledgments}
Bingzhe Wu and Guangyu Sun are supported by National Natural Science Foundation of China (No.61572045).
\bibliographystyle{aaai}
\bibliography{ref}

\begin{thebibliography}{}

\bibitem[\protect\citeauthoryear{Brodersen \bgroup et al\mbox.\egroup
  }{2010}]{brodersen2010balanced}
Brodersen, K.; Ong, C.; Stephan, K.; and Buhmann, J.
\newblock 2010.
\newblock The balanced accuracy and its posterior distribution.
\newblock In {\em ICPR}.

\bibitem[\protect\citeauthoryear{Chen \bgroup et al\mbox.\egroup
  }{2016}]{chen2016mitosis}
Chen, H.; Dou, Q.; Wang, X.; Qin, J.; Heng, P.; et~al.
\newblock 2016.
\newblock Mitosis detection in breast cancer histology images via deep cascaded
  networks.
\newblock In {\em AAAI}.

\bibitem[\protect\citeauthoryear{Cruz-Roa \bgroup et al\mbox.\egroup
  }{2014}]{cruz2014automatic}
Cruz-Roa, A.; Basavanhally, A.; Gonz{\'a}lez, F.; Gilmore, H.; Feldman, M.;
  et~al.
\newblock 2014.
\newblock Automatic detection of invasive ductal carcinoma in whole slide
  images with convolutional neural networks.
\newblock In {\em SPIE}.

\bibitem[\protect\citeauthoryear{Dou \bgroup et al\mbox.\egroup
  }{2016}]{Dou2016}
Dou, Q.; Chen, H.; Yu, L.; Zhao, L.; Qin, J.; et~al.
\newblock 2016.
\newblock Automatic detection of cerebral microbleeds from mr images via 3d
  convolutional neural networks.
\newblock {\em TMI} 35(5):1182--1195.

\bibitem[\protect\citeauthoryear{Esteva \bgroup et al\mbox.\egroup
  }{2017}]{Esteva2017}
Esteva, A.; Kuprel, B.; Novoa, R.; Ko, J.; Swetter, S.; et~al.
\newblock 2017.
\newblock Dermatologist-level classification of skin cancer with deep neural
  networks.
\newblock {\em Nature} 542(7639):115--118.

\bibitem[\protect\citeauthoryear{Fogo \bgroup et al\mbox.\egroup
  }{2014}]{fogo2014fundamentals}
Fogo, A.; Cohen, A.; Colvin, R.; Jennette, J.~C.; and Alpers, C.
\newblock 2014.
\newblock {\em Fundamentals of Renal Pathology}.
\newblock Springer.

\bibitem[\protect\citeauthoryear{Gallego \bgroup et al\mbox.\egroup
  }{2018}]{Gallego2018Glomerulus}
Gallego, J.; Pedraza, A.; Lopez, S.; Steiner, G.; Gonzalez, L.; Laurinavicius,
  A.; and Bueno, G.
\newblock 2018.
\newblock Glomerulus classification and detection based on convolutional neural
  networks.
\newblock {\em Journal of Imaging} 4(1).

\bibitem[\protect\citeauthoryear{Goodfellow \bgroup et al\mbox.\egroup
  }{2014}]{goodfellow2014generative}
Goodfellow, I.; Pouget-Abadie, J.; Mirza, M.; Xu, B.; Warde-Farley, D.; Ozair,
  S.; Courville, A.; and Bengio, Y.
\newblock 2014.
\newblock Generative adversarial nets.
\newblock In {\em NIPS}.

\bibitem[\protect\citeauthoryear{Gulshan \bgroup et al\mbox.\egroup
  }{2016}]{Gulshan2016}
Gulshan, V.; Peng, L.; Coram, M.; Stumpe, M.; Wu, D.; et~al.
\newblock 2016.
\newblock Development and validation of a deep learning algorithm for detection
  of diabetic retinopathy in retinal fundus photographs.
\newblock {\em JAMA} 304(6):649--656.

\bibitem[\protect\citeauthoryear{Han and Ma}{2002}]{han2002fuzzy}
Han, J., and Ma, K.
\newblock 2002.
\newblock Fuzzy color histogram and its use in color image retrieval.
\newblock {\em TIP} 11(8):944--952.

\bibitem[\protect\citeauthoryear{He \bgroup et al\mbox.\egroup
  }{2016}]{he2016deep}
He, K.; Zhang, X.; Ren, S.; and Sun, J.
\newblock 2016.
\newblock Deep residual learning for image recognition.
\newblock In {\em CVPR}.

\bibitem[\protect\citeauthoryear{Hu, Shen, and Sun}{2018}]{hu2018senet}
Hu, J.; Shen, L.; and Sun, G.
\newblock 2018.
\newblock Squeeze-and-excitation networks.
\newblock In {\em CVPR}.

\bibitem[\protect\citeauthoryear{Isola \bgroup et al\mbox.\egroup
  }{2017}]{isola2016image}
Isola, P.; Zhu, J.; Zhou, T.; and Efros, A.
\newblock 2017.
\newblock Image-to-image translation with conditional adversarial networks.
\newblock In {\em CVPR}.

\bibitem[\protect\citeauthoryear{Janowczyk and
  Madabhushi}{2016}]{Janowczyk2016Deep}
Janowczyk, A., and Madabhushi, A.
\newblock 2016.
\newblock Deep learning for digital pathology image analysis: A comprehensive
  tutorial with selected use cases.
\newblock {\em JPI} 7(1):29.

\bibitem[\protect\citeauthoryear{Kakimoto \bgroup et al\mbox.\egroup
  }{2014}]{Kakimoto2014Automated}
Kakimoto, T.; Okada, K.; Hirohashi, Y.; Relator, R.; Kawai, M.; et~al.
\newblock 2014.
\newblock Automated image analysis of a glomerular injury marker desmin in
  spontaneously diabetic torii rats treated with losartan.
\newblock {\em Journal of Endocrinology} 222(1):43--51.

\bibitem[\protect\citeauthoryear{Kim \bgroup et al\mbox.\egroup
  }{2017}]{kim2017learning}
Kim, T.; Cha, M.; Kim, H.; Lee, J.; and Kim, J.
\newblock 2017.
\newblock Learning to discover cross-domain relations with generative
  adversarial networks.
\newblock In {\em ICML}.

\bibitem[\protect\citeauthoryear{Krizhevsky, Sutskever, and
  Hinton}{2012}]{krizhevsky2012imagenet}
Krizhevsky, A.; Sutskever, I.; and Hinton, G.
\newblock 2012.
\newblock Imagenet classification with deep convolutional neural networks.
\newblock In {\em NIPS}.

\bibitem[\protect\citeauthoryear{Levin \bgroup et al\mbox.\egroup
  }{2017}]{Lancet2017}
Levin, A.; Tonelli, M.; Bonventre, J.; Coresh, J.; Donner, J.; Fogo, A.; et~al.
\newblock 2017.
\newblock Global kidney health 2017 and beyond: A roadmap for closing gaps in
  care, research, and policy.
\newblock {\em The Lancet} 390(10105):1888--1917.

\bibitem[\protect\citeauthoryear{Lin \bgroup et al\mbox.\egroup
  }{2017}]{lin2017focal}
Lin, T.; Goyal, P.; Girshick, R.; He, K.; and Doll{\'a}r, P.
\newblock 2017.
\newblock Focal loss for dense object detection.
\newblock In {\em ICCV}.

\bibitem[\protect\citeauthoryear{Liu \bgroup et al\mbox.\egroup
  }{2017}]{liu2017detecting}
Liu, Y.; Gadepalli, K.; Norouzi, M.; Dahl, G.; Kohlberger, T.; Boyko, A.;
  Venugopalan, S.; Timofeev, A.; Nelson, P.; Corrado, G.; et~al.
\newblock 2017.
\newblock Detecting cancer metastases on gigapixel pathology images.
\newblock In {\em CoRR}.

\bibitem[\protect\citeauthoryear{Liu, Breuel, and
  Kautz}{2017}]{liu2017unsupervised}
Liu, M.; Breuel, T.; and Kautz, J.
\newblock 2017.
\newblock Unsupervised image-to-image translation networks.
\newblock In {\em NIPS}.

\bibitem[\protect\citeauthoryear{Pedraza \bgroup et al\mbox.\egroup
  }{2017}]{Pedraza2017Glomerulus}
Pedraza, A.; Gallego, J.; Lopez, S.; Gonzalez, L.; Laurinavicius, A.; and
  Bueno, G.
\newblock 2017.
\newblock Glomerulus classification with convolutional neural networks.
\newblock In {\em MIUA}.

\bibitem[\protect\citeauthoryear{Ronneberger, Fischer, and
  Brox}{2015}]{ronneberger2015u}
Ronneberger, O.; Fischer, P.; and Brox, T.
\newblock 2015.
\newblock U-net: Convolutional networks for biomedical image segmentation.
\newblock In {\em MICCAI}.

\bibitem[\protect\citeauthoryear{Shrivastava \bgroup et al\mbox.\egroup
  }{2017}]{shrivastava2017learning}
Shrivastava, A.; Pfister, T.; Tuzel, O.; Susskind, J.; Wang, W.; and Webb, R.
\newblock 2017.
\newblock Learning from simulated and unsupervised images through adversarial
  training.
\newblock In {\em CVPR}.

\bibitem[\protect\citeauthoryear{Simonyan and
  Zisserman}{2015}]{simonyan2014very}
Simonyan, K., and Zisserman, A.
\newblock 2015.
\newblock Very deep convolutional networks for large-scale image recognition.
\newblock In {\em ICLR}.

\bibitem[\protect\citeauthoryear{Szegedy \bgroup et al\mbox.\egroup
  }{2017}]{szegedy2017inception}
Szegedy, C.; Ioffe, S.; Vanhoucke, V.; and Alemi, A.
\newblock 2017.
\newblock Inception-v4, inception-resnet and the impact of residual connections
  on learning.
\newblock In {\em AAAI}.

\bibitem[\protect\citeauthoryear{Yi \bgroup et al\mbox.\egroup
  }{2017}]{yi2017dualgan}
Yi, Z.; Zhang, H.; Gong, P.; et~al.
\newblock 2017.
\newblock Dualgan: Unsupervised dual learning for image-to-image translation.
\newblock In {\em ICCV}.

\bibitem[\protect\citeauthoryear{Zhu \bgroup et al\mbox.\egroup
  }{2017}]{zhu2017unpaired}
Zhu, J.; Park, T.; Isola, P.; and Efros, A.
\newblock 2017.
\newblock Unpaired image-to-image translation using cycle-consistent
  adversarial networks.
\newblock In {\em ICCV}.

\end{thebibliography}
\end{document}